%% file: main.tex
\pdfoutput=1
\documentclass[sigconf, nonacm]{acmart}
\usepackage{booktabs} 
\usepackage{bm}
\usepackage{subcaption}
\usepackage{algorithm}
\usepackage{algpseudocode}
\usepackage{hyperref}

\graphicspath{{figures/}}

\algdef{SE}[SUBALG]{Indent}{EndIndent}{}{\algorithmicend\ }%
\algtext*{Indent}
\algtext*{EndIndent}

\copyrightyear{2021}
\acmYear{2021}
\setcopyright{acmcopyright}\acmConference[GECCO '21 Companion]{2021 Genetic
and Evolutionary Computation Conference Companion}{July 10--14, 2021}{Lille,
France}
\acmBooktitle{2021 Genetic and Evolutionary Computation Conference Companion
(GECCO '21 Companion), July 10--14, 2021, Lille, France}
\acmPrice{15.00}
\acmDOI{10.1145/3449726.3463214}
\acmISBN{978-1-4503-8351-6/21/07}

\renewcommand\footnotetextcopyrightpermission[1]{}
\begin{document}
\title[Quantifying the Impact of Boundary Constraint Handling Methods on DE]{Quantifying the Impact of Boundary Constraint Handling Methods on Differential Evolution}

\author{Rick Boks}
\affiliation{%
  \institution{LIACS, Leiden University}
  \country{The Netherlands}
}
\email{rick@rickboks.com}

\author{Anna V. Kononova}
\orcid{0002-4138-7024}
\affiliation{%
  \institution{LIACS, Leiden University}
  \country{The Netherlands}
}
\email{a.kononova@liacs.leidenuniv.nl}

\author{Hao Wang}
\affiliation{%
  \institution{LIACS, Leiden University}
  \country{The Netherlands}
}
\email{h.wang@liacs.leidenuniv.nl}

\renewcommand{\shortauthors}{R. Boks et al.}

\begin{abstract}
Constraint handling is one of the most influential aspects of applying metaheuristics to real-world applications, which can hamper the search progress if treated improperly. In this work, we focus on a particular case - the box constraints, for which many boundary constraint handling methods (BCHMs) have been proposed. We call for the necessity of studying the impact of BCHMs on metaheuristics' performance and behavior, which receives seemingly little attention in the field. We target quantifying such impacts through systematic benchmarking by investigating 28 major variants of Differential Evolution (DE) taken from the modular DE framework (by combining different mutation and crossover operators) and $13$ commonly applied BCHMs, resulting in $28 \times 13 = 364$ algorithm instances after pairing DE variants with BCHMs. After executing the algorithm instances on the well-known BBOB/COCO problem set, we analyze the best-reached objective function value (performance-wise) and the percentage of repaired solutions (behavioral) using statistical ranking methods for each combination of mutation, crossover, and BBOB function group. Our results clearly show that the choice of BCHMs substantially affects the empirical performance as well as the number of generated infeasible solutions, which allows us to provide general guidelines for selecting an appropriate BCHM for a given scenario.
\end{abstract}

%
%


\maketitle

\input{introduction}
\input{de}
\input{mutation}
\input{crossover}
\input{parameters}
\input{bchm}
\input{experiments}
\input{results}
\begin{figure*}
\centering
\includegraphics[width=.94\textwidth]{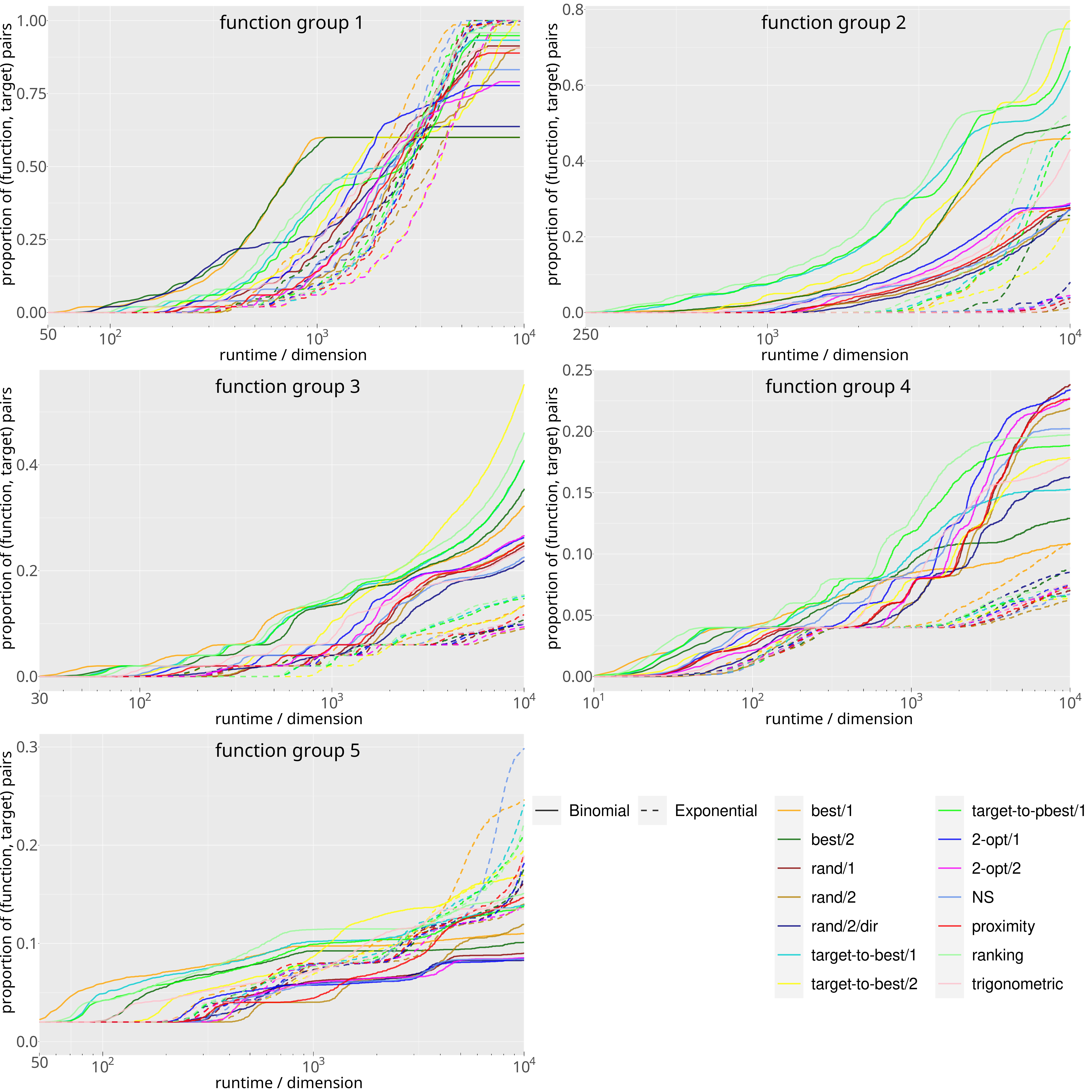}
\vspace{-0.3cm}
\caption{
Empirical Cumulative Distribution Functions (ECDFs) of the best DE instances for each combination of mutation and crossover (with the best BCHM according to Figure~\ref{fig:heatmaps}), aggregated over the 100 runs of each function in each function group. Note that the plotted DE instances differ across the 5 function groups.}\label{fig:combined_ecdf}
\vspace{-0.4cm}
\end{figure*}
\input{conclusion}

\bibliographystyle{ACM-Reference-Format}
\bibliography{bibliography} 

\end{document}

%% file: introduction.tex
\section{Introduction} \label{sec:introduction}
Since its debut in 1995~\cite{firstde}, Differential Evolution (DE) has been developed into one of the well-known optimization algorithms for solving non-linear continuous optimization problems~\cite{Plagianakos2008,BilalPZGA20}. Due to its simplicity and robustness, DE has been applied to various real-world applications, e.g., data mining~\cite{AlatasAK08}, neural network training~\cite{IlonenKL03}, and hyper-parameter tuning~\cite{abs-2012-08180}. As its most distinguishing feature, the mutation operator of DE is designed to take as input the difference between some randomly chosen individuals (called a differential vector) in the population. Despite the elegance of this mutation operator, it might become problematic when solving a constrained problem (e.g., objective functions with box constraints). When the constraints are violated by some individuals (which might happen due to randomness), it is likely to produce a huge differential vector, which will yield more violations when used to create new individuals. Our arguments are seconded in~\cite{outside-the-box}, which shows that during an optimization run of DE, a large number of infeasible
solutions can be generated, where the percentage of generated
infeasible solutions increases for problems with higher dimensionality. 
We therefore call for the importance of careful treatment and consideration of constraints when applying a DE algorithm. Particularly, in this paper, we initiate our study of the impact of boundary constraint handling methods (BCHMs) by investigating closely into a special case - the box constraints.

Notably, in~\cite{outside-the-box}, it has been shown that on the noisy landscape $f_0$,
where $\forall \bm{x} \in \mathbb{R}^n : f_0(\bm{x}) \sim \mathcal{U}(0,1)$, most DE configurations (e.g., DE/rand/1/bin) with commonly used settings for the hyper-parameters (e.g., $F$ and $Cr$) yield infeasible solutions almost exclusively (with a probability of nearly $100\%$), when coupled with some boundary constraint handling methods. Despite its simplicity, $f_0$ can serve as an ideal testbed for justifying a BCHM for DEs, since it is scalable, and more importantly, imposes no selection pressure over the search, on which DE will violate the constraints quite often and break each constraint boundary with the same probability.
We found it worrisome that 1) the choice of BCHM is rarely given the appropriate amount of attention~\cite{rafal, anna-bias} and 2) BCHMs can significantly impact the overall performance of the algorithm~\cite{arabas} and the degree of structural bias~\cite{anna-bias}.

To address this issue, we have contrived an empirical approach, which incorporates 13 well-known BCHMs (e.g., resampling, wrapping, and boundary transformation) and then tested each BCHM with a wide range of DE variants~\cite{PSODE}, e.g., combinations of mutation and crossover strategies, on the well-known BBOB/COCO problem set~\cite{coco}, for answering the research question of whether a certain combination of a DE variant and a BCHM behaves very differently compared to other combinations, on (a subset of) the BBOB problem set. We will analyze the behavioral impact of BCHMs in terms of empirical performance and the proportion of infeasible solutions created during the run.

This paper is organized as follows: Section~\ref{sec:de} covers the basic
structure of DE, variants of its mutation and crossover operators, and the employed parameter adaptation scheme;
Section~\ref{sec:bchm} describes a wide selection of BCHMs which can be applied to DE;
Section~\ref{sec:experiments} discusses our experiments; in Section~\ref{sec:results}, we analyze
the experiment results; finally, in Section~\ref{sec:conclusion} we summarize the main findings of
this paper and give directions for future work.

%% file: de.tex
\section{Differential Evolution} \label{sec:de}
Formally, in this work, we deal with a single-objective function supported on a closed subset of $\mathbb{R}^n$, i.e., $f\colon D \subset \mathbb{R}^n \rightarrow \mathbb{R}$, where $D = [x^{\text{min}}_1, x^{\text{max}}_1] \times  \cdots \times [x^{\text{min}}_n, x^{\text{max}}_n]$. We start with delineating DE's working mechanism (see Alg.~\ref{alg:de} for the pseudo-code). DE initializes  the population of individuals $\{\mathbf{x}_1,\mathbf{x}_2,\ldots, \mathbf{x}_M\}\subset\mathbb{R}^n$ (with $M$ being its size) by sampling each individual uniformly at random (u.a.r.) in $D$. Also, in the following discussions, we shall represent the $j$th component of vector $\bm{x}_i$ by $x_{i,j}$. 

For each individual $\bm{x}_i$, a \textit{donor vector} $\bm{v}_i$
(a.k.a.~mutant) is generated through mutation, a process where scaled difference vectors are added
to a \emph{base vector}. Please see Section~\ref{sec:mutation} for an overview of some commonly used mutation strategies. Subsequently, for each donor vector
$\bm{v}_{i}$,  a \textit{trial vector} $\bm{x}_i'$ is created by means of crossover, where
components are exchanged between the parent vector and the donor vector. Two crossover methods are
most prominent in DE literature, both of which are described in Section~\ref{sec:crossover}.
Elitist selection is applied between $\bm{x}_i$ and $\bm{x}_i'$, where the better one is kept for
the next iteration.

\begin{algorithm}[!ht]
\begin{algorithmic}[1]
	\State{$\bm{x}_i \leftarrow \bm{\mathcal{U}}(\bm{x}^{\text{min}}, \bm{x}^{\text{max}}), \quad i =1, \ldots, M.$}\Comment{Initialize}
	\While{termination criteria are not met}
		\For{$i = 1 \rightarrow M$}
			\State{Choose $r_1 \neq r_2 \neq r_3 \neq i \in [1..M]$ u.a.r.}
			\State{$\bm{v}_i \leftarrow \bm{x}_{r_{1}} + F \cdot (\bm{x}_{r_{2}} - \bm{x}_{r_{3}})$}\Comment{Mutate}
			\State{Choose $j_{\text{rand}} \in [1..n]$ u.a.r.}
			\For {$j = 1 \rightarrow n$}
    			\If{$\mathcal{U}(0,1) \leq Cr$ or $j = j_{\text{rand}}$}
				\State $x_{i,j}' \leftarrow v_{i,j}$ \Comment{Crossover}
                \Else
                    \State $x_{i,j}' \leftarrow x_{i,j}$
                \EndIf
            \EndFor
		\EndFor
		\For{$i = 1 \rightarrow M$}
            \If{$f(\bm{x}_i') < f(\bm{x}_i)$}
				\State $\bm{x}_i \leftarrow \bm{x}_i'$ \Comment{Select}	
			\EndIf
		\EndFor
	\EndWhile
\caption{Rand/1/bin Differential Evolution}
\label{alg:de}
\end{algorithmic}
\end{algorithm}

%% file: mutation.tex
\subsection{Mutation} \label{sec:mutation}
Here, we discuss the $14$ mutation schemes considered in our
experiments. Typically, vectors with indices $r_1, r_2, \dots, r_j$ are selected uniformly at
random without replacement, where $\forall j: r_j \neq i$. Depending on the chosen
mutation scheme, these randomly selected vectors can appear both in the difference vectors and as
the base vector. The difference vectors are scaled by the so-called \emph{mutation rate} $F > 0$, 
which controls the strength of the mutation operator. Although there is no upper limit, $F > 1$ is
rarely considered effective~\cite{de-book}. 

In addition to the five well-known mutation operators from the original DE `family' by Storn and Price: rand/1, best/1, target-to-best/1, best/2, and rand/2, we consider nine other prominent variants proposed in  literature in an attempt to enhance DE
performance:

\begin{itemize}

\item \textbf{Target-to-best/2}~\cite{ranking} \label{par:target_to_best_2}
\begin{equation}
   \begin{aligned}
  \bm{v}_i \leftarrow \bm{x}_{i} + F \cdot (\bm{x}_{\text{best}} - 
  \bm{x}_{i}) + F \cdot (\bm{x}_{r_{1}} - \bm{x}_{r_{2}})
  \\ + F \cdot (\bm{x}_{r_{3}} - \bm{x}_{r_{4}}),
\end{aligned} 
\end{equation} where $\bm{x}_{\text{best}}$ is the best member of the population.

\item \textbf{Target-to-$p$best/1}~\cite{jade} \label{par:target_to_p_best_1}

\begin{equation} \label{eq:pbest}
  \bm{v}_i \leftarrow \bm{x}_{i} + F \cdot (\bm{x}_{\text{best}}^p - 
  \bm{x}_{i}) + F \cdot (\bm{x}_{r_{1}} - \bm{x}_{r_{2}}),
\end{equation}
where $\bm{x}_{best}^{p}$ is selected uniformly at random from the best $p \cdot 100\%$ members of the
population, $p \in (0,1]$.  We set $p = \max(0.05, 3/M)$, as recommended in~\cite{jade}.

\item \textbf{Rand/2/dir}~\cite{comparative} \label{par:rand_2_dir}
\begin{equation}
\begin{aligned}
  \bm{v}_{i} \leftarrow \bm{x}_{r_{1}} + \frac{F}{2} \cdot 
  ( \bm{x}_{r_1} - \bm{x}_{r_2} + \bm{x}_{r_3} - \bm{x}_{r_4} ), 
\end{aligned}
\end{equation}
where $f( \bm{x}_{r_1} ) < f ( \bm{x}_{r_2} )$ and $f( \bm{x}_{r_3} ) < f( \bm{x}_{r_4} )$.

\item \textbf{NSDE}~\cite{NSDE} \label{par:nsde}
\begin{equation}
\begin{aligned}
  \bm{v}_{i} \leftarrow \bm{x}_{r_1} + ( \bm{x}_{r_2} - \bm{x}_{r_3} ) \cdot \begin{cases}
      \mathcal{N}(0.5,0.5) & \\ \text{ \qquad if } \mathcal{U}(0,1) < 0.5 \\
      \mathcal{C}(0,1) & \\ \text{ \qquad otherwise}
\end{cases},
\end{aligned}
\end{equation}
where $\mathcal{N}$ is a normal distribution, $\mathcal{C}$ a Cauchy distribution, and $\mathcal{U}$
a uniform distribution.
\item \textbf{Trigonometric}~\cite{trigonometric} \label{par:trigonometric}
\begin{equation}
\begin{aligned}
  \bm{v}_{i} \leftarrow (\bm{x}_{r_1} + \bm{x}_{r_2} + \bm{x}_{r_3})/3 + (p_2 - p_1) \cdot 
  (\bm{x}_{r_1} - \bm{x}_{r_2})+ \\ (p_3 - p_2)
  \cdot (\bm{x}_{r_2} - \bm{x}_{r_3}) + (p_1 - p_3) \cdot (\bm{x}_{r_3} - \bm{x}_{r_1}),
\end{aligned}
\end{equation}
where\:\:$p_1 = \lvert f( \bm{x}_{r_1} ) \rvert / p'$,\:\:$p_2 = \lvert f( \bm{x}_{r_2} ) \rvert /
p'$,\:\:$p_3 = \lvert f( \bm{x}_{r_3} ) \rvert / p'$
and\:\:$p' = \lvert f( \bm{x}_{r_1} ) \rvert + \lvert f( \bm{x}_{r_2} ) \rvert + \lvert f( \bm{x}_{r_3} ) \rvert$.
This mutation scheme is applied with probability $\Gamma = 0.05$, otherwise rand/1 mutation is applied.
\item \textbf{2-Opt/1}~\cite{twoopt} \label{par:two_opt_1}
\begin{equation} \label{eq:two-opt1} 
 \bm{v}_i \leftarrow 
  \begin{cases}
    \bm{x}_{r_1} + F \cdot (\bm{x}_{r_2} - \bm{x}_{r_3}) & \text{ if } f(\bm{x}_{r_1}) < f(\bm{x}_{r_2}) \\
    \bm{x}_{r_2} + F \cdot (\bm{x}_{r_1} - \bm{x}_{r_3}) & \text{ otherwise}
  \end{cases}
\end{equation}

\item \textbf{2-Opt/2}~\cite{twoopt} \label{par:two_opt_2}

\begin{equation}
\begin{aligned}
 \bm{v}_i \leftarrow
  \begin{cases}
   \bm{x}_{r_1} + F \cdot (\bm{x}_{r_2} - \bm{x}_{r_3}) + 
   F \cdot (\bm{x}_{r_4} - \bm{x}_{r_5}) & \\ \text{ \qquad if } f(\bm{x}_{r_1}) < f(\bm{x}_{r_2}) \\
    \bm{x}_{r_2} + F \cdot (\bm{x}_{r_1} - \bm{x}_{r_3}) + 
   F \cdot (\bm{x}_{r_4} - \bm{x}_{r_5}) & \\ \text{ \qquad otherwise}
   \end{cases}
   \end{aligned}
  \end{equation}


\item \textbf{Proximity-based rand/1}~\cite{proximity} \label{par:proximity_based_rand_1}
Here, the indices $r_j$ are chosen by a roulette wheel without replacement, where the selection
probability of each index is proportional to the Euclidean distance from the target vector to the
corresponding individual. The authors~\cite{proximity} reported the most significant performance
improvement in conjunction with exploratory mutation schemes. Therefore, we choose rand/1 mutation
for our experiments with the proximity-based approach.

\item \textbf{Ranking-based target-to-$p$best/1}~\cite{ranking} \label{par:ranking_based_target_to_p_best_1}
Target-to-$p$best/1 (Eq.~\ref{eq:pbest}) is used, and the index $r_1$ is selected using the roulette
wheel method. The selection probability of an individual is proportional to its rank in the
population w.r.t. the fitness values. 
\end{itemize}

%% file: crossover.tex
\subsection{Crossover} \label{sec:crossover}
The crossover step in DE exchanges elements between the target vector and the donor vector (resulting
from the mutation step). The resulting vector is called the \emph{trial} vector. We consider the two most commonly used crossover schemes: binomial and exponential crossover.

\paragraph{Binomial Crossover~\cite{firstde}} One component, which is selected u.a.r. in $[1,n]$, is always inherited from the donor vector. Each remaining component is inherited from the donor vector with probability $Cr\in[0, 1]$, and copied from the target vector otherwise.

\paragraph{Exponential Crossover~\cite{firstde}} A starting index in $[1,n]$ is selected u.a.r., and consecutive components (using wrapping) are inherited from the donor vector until the condition $\mathcal{U} (0,1) < Cr$ is violated (or all components are already inherited), and the exchange of components stops. The remaining values are copied from the target vector.

%% file: parameters.tex
\subsection{Adaptation of Control Parameters} \label{sec:self_adaptation}
An ongoing problem in the field of evolutionary algorithms is the sensitivity to control parameters.
Differential Evolution has relatively few parameters, namely the mutation rate $F$, crossover rate
$Cr$, and population size $M$.  Still, as the optimal settings of control parameters is
problem-dependent~\cite{firstde, parameterstudy, comparative, realparamde}, tuning these
parameters is essential in order to obtain the desired result. 

For this reason, much effort has gone toward adapting the parameter values during the
optimization process, for example in jDE~\cite{comparative-sa}, JADE~\cite{jade}, and
SaDE~\cite{sade}. In our experiments, we use the state-of-the-art scheme for adaptation of $F$ and $Cr$ proposed as part of SHADE~\cite{shade}.

%% file: bchm.tex
\section{Boundary Constraint Handling Methods} \label{sec:bchm}
In this work, we consider $13$ BCHMs which, unless stated otherwise, are
applied directly after the mutation step. The reason for this is that, when dealing with box constraints, the
crossover operator in DE is feasibility preserving, i.e., the crossover product of two feasible
solution vectors is always another feasible solution vector. Moreover, such setup allows easier tracking of percentage of repaired solutions (see Section~\ref{sub:comparison_of_bchms}).
The repair methods considered in this
work all follow the Lamarckian evolution model: an infeasible individual is replaced by its repaired
version, where each component with index $j$ is positioned between the corresponding lower 
bound $x^{\text{min}}_j$, and upper bound $x^{\text{max}}_j$ of the feasible space.
In the Darwinian evolution model, the position of an infeasible individual is not altered by the
BCHM, but it is assigned the fitness value of its repaired version, thereby maintaining infeasible
solutions in the population. We do not consider BCHMs of the latter type.

\paragraph{Resampling} \label{par:resampling}
In the resampling method~\cite{arabas}, when an infeasible solution is generated, the mutation
operator is re-applied to the entire solution vector until a feasible solution is obtained. Phrased
differently, the selection of the random indices $r_i$ is repeated until the result of the mutation
is feasible. We set the maximum number of resamples to $100$, after which the solution is repaired
by means of projection (discussed later).

\paragraph{Random Reinitialization~\cite{de-book}} \label{par:reinitialization} 
Infeasible components are reinitialized randomly inside the bounds of the search space. 
\begin{equation}
  v_{i,j} \leftarrow \begin{cases}
    \mathcal{U}(x^{\text{min}}_j, x^{\text{max}}_j) & \text{if } v_{i,j} < x^{\text{min}}_j \text{ or } v_{i,j} > x^{\text{max}}_j \\
    v_{i,j} & \text{otherwise} 
  \end{cases}
\end{equation}

\paragraph{Projection~\cite{comparative-sa}} \label{par:projection}
Infeasible components are placed on the violated boundary:
\begin{equation}
  v_{i,j} \leftarrow \begin{cases}
    x^{\text{min}}_j & \text{if } v_{i,j} < x^{\text{min}}_j \\
    x^{\text{max}}_j & \text{if } v_{i,j} > x^{\text{max}}_j \\
    v_{i,j} & \text{otherwise} 
  \end{cases}
\end{equation}

\paragraph{Reflection~\cite{realparamde}} \label{par:reflection}
Infeasible components are reflected to the other side of the violated boundary. This can result in
the component being placed outside of the opposite boundary, in which case the reflection is
repeated until a feasible component is obtained.

\begin{equation}
  v_{i,j} \leftarrow \begin{cases}
    2x^{\text{min}}_j - v_{i,j} & \text{if } v_{i,j} < x^{\text{min}}_j \\
    2x^{\text{max}}_j - v_{i,j} & \text{if } v_{i,j} > x^{\text{max}}_j \\
    v_{i,j} & \text{otherwise} 
  \end{cases}
\end{equation}

\paragraph{Wrapping~\cite{wrapping}} \label{par:wrapping}
This method assumes the search space to be of a toroidal shape, making infeasible components enter
the search space on the opposite side.
\begin{equation}
  v_{i,j} \leftarrow \begin{cases}
    x^{\text{max}}_j - (x^{\text{min}}_j - v_{i,j})(\text{mod } \lvert x^{\text{max}}_j - x^{\text{min}}_j \rvert) & \text{if } v_{i,j} < x^{\text{min}}_j \\
    x^{\text{min}}_j + (v_{i,j} - x^{\text{max}}_j)(\text{mod } \lvert x^{\text{max}}_j-x^{\text{min}}_j \rvert  )& \text{if } v_{i,j} > x^{\text{max}}_j \\
    v_{i,j} & \text{otherwise} 
  \end{cases}
\end{equation}

\paragraph{Boundary Transformation~\cite{c-cmaes}} \label{par:transformation} This BCHM is used by default in 
the implementation of the CMA-ES in
Python~\cite{cmaes-python} and C~\cite{c-cmaes}. In contrast to other BCHMs discussed in this work, it \textit{can} also perturb
feasible individuals. Predetermined
offsets $\bm{a}^{l}$ and $\bm{a}^{u}$ from the lower and upper bounds are used: 
\begin{align}
  a^l_j = \min( (x^{\text{max}}_j - x^{\text{min}}_j)/2, 1 + \lvert x^{\text{min}}_j\rvert/20 ) \\
  a^u_j = \min( (x^{\text{max}}_j - x^{\text{min}}_j)/2, 1 + \lvert x^{\text{max}}_j\rvert/20 )
\end{align}
Values $v_{i,j} \in [x^{\text{min}}_j + a^l_j, x^{\text{max}}_j - a^u_j]$ are not modified. Values in
$[x^{\text{min}}_j - 3a^l_j, x^{\text{min}}_j - a^l_j]$ and $[x^{\text{max}}_j + a^u_j, x^{\text{max}}_j + 3a^u_j]$ are
first shifted into the feasible preimage $[x^{\text{min}}_j - a^l_j, x^{\text{max}}_j + a^u_j]$ by reflecting the
value using $x^{\text{min}}_j - a^l_j$ or $x^{\text{max}}_j + a^u_j$ respectively as a bound. Values further away from
the boundaries are shifted upwards or downwards by a periodic transformation with period $2 \cdot (x^{\text{max}}_j -
x^{\text{min}}_j + a^l_j + a^u_j)$. After the component is shifted into the feasible preimage, it is transformed as follows:

\begin{equation}
  v_{i,j} \leftarrow \begin{cases}
    x^{\text{min}}_j + (v_{i,j} - (x^{\text{min}}_j - a^l_j))^2 / 4a^l_j & \text{if } v_{i,j} <
    x^{\text{min}}_j + a^l_j \\
    x^{\text{max}}_j - (v_{i,j} - (x^{\text{max}}_j + a^u_j))^2 / 4a^u_j & \text{if } v_{i,j} >
    x^{\text{max}}_j - a^u_j \\
    v_{i,j} & \text{otherwise}\\
  \end{cases} 
\end{equation}

\paragraph{Rand Base~\cite{de-book}} \label{par:rand_base}
This method places infeasible components on a random location between the violated boundary and
the corresponding component of the base vector $\bm{b}$.
\begin{equation}
  v_{i,j} \leftarrow \begin{cases}
    \mathcal{U}(x^{\text{min}}_j, b_j) & \text{if } v_{i,j} < x^{\text{min}}_j \\
    \mathcal{U}(b_j, x^{\text{max}}_j) & \text{if } v_{i,j} > x^{\text{max}}_j \\
    v_{i,j} & \text{otherwise}
  \end{cases}
\end{equation}

\paragraph{Midpoint Base~\cite{de-book}} \label{par:midpoint_base}
This method places infeasible components halfway between the violated boundary and the corresponding component
of the base vector $\bm{b}$.
\begin{equation}
  v_{i,j} \leftarrow \begin{cases}
    (x^{\text{min}}_j + b_j) / 2 & \text{if } v_{i,j} < x^{\text{min}}_j \\
    (b_j + x^{\text{max}}_j) / 2 & \text{if } v_{i,j} > x^{\text{max}}_j \\
    v_{i,j} & \text{otherwise}
  \end{cases}
\end{equation}

\paragraph{Midpoint Target~\cite{rafal}} \label{par:midpoint_target}
This method places infeasible components halfway between the violated boundary and the corresponding
component of the target vector $\bm{t}$.
\begin{equation}
  v_{i,j} \leftarrow \begin{cases}
    (x^{\text{min}}_j + t_j) / 2 & \text{if } v_{i,j} < x^{\text{min}}_j \\
    (t_j + x^{\text{max}}_j) / 2 & \text{if } v_{i,j} > x^{\text{max}}_j \\
    v_{i,j} & \text{otherwise}
  \end{cases}
\end{equation}

\paragraph{Conservatism~\cite{arabas}} \label{par:conservatism}
If the individual is infeasible, i.e., at least one of its components is infeasible, the
\emph{entire vector} is copied from the base vector $\bm{b}$.

\begin{equation}
  \bm{v}_{i} \leftarrow \begin{cases}
    \bm{b} & \text{if } \bm{v}_{i} \not\in D \\ 
    \bm{v}_{i}  & \text{otherwise}
  \end{cases}
\end{equation}

\paragraph{Projection to Midpoint~\cite{kreischer}} \label{par:projection_to_midpoint}
This method projects an infeasible individual onto the boundary of the search space, towards the 
center of the search space:

\begin{equation}
  \bm{v}_{i} \leftarrow (1- \alpha) \cdot (\bm{x}^{\text{min}} + \bm{x}^{\text{max}})/2 + \alpha \cdot \bm{v}_{i}, 
\end{equation}
where $\alpha \in [0,1]$ is the largest value such that $x^{\text{min}}_j \leq v_{i,j} \leq x^{\text{max}}_j$, for
all $j \in \{1, \dots, n\}$.

\paragraph{Projection to Base~\cite{rafal}} \label{par:projection_to_base}
Similar to `projection to midpoint', but the projection is performed towards the base vector $\bm{b}$: 
\begin{equation}
  \bm{v}_{i} \leftarrow (1-\alpha) \cdot \bm{b} + \alpha \cdot \bm{v}_i 
\end{equation}
where $\alpha \in [0,1]$ is the largest value such that $x^{\text{min}}_j \leq v_{i,j} \leq x^{\text{max}}_j$, for
all $j \in \{1, \dots, n\}$.

\paragraph{Death Penalty~\cite{de-book}} \label{par:death_penalty}
An infeasible individual is assigned an arbitrarily large fitness value, greater than any individual
in the feasible space can obtain. This will ensure that the resulting trial vector is not accepted
in the selection step. This BCHM is applied directly after the crossover step.

%% file: experiments.tex
\section{Experiments} \label{sec:experiments}
A modular DE framework\footnote{The source code is available at:~\url{https://github.com/rickboks/pso-de-framework}} 
was implemented in C\texttt{++}, in which
each previously discussed mutation method, crossover method, and BCHM can be combined arbitrarily.
We consider~$14$ mutation methods, $2$ crossover methods and $13$ BCHMs, using which we can generate
a total of~$14 \times  2 \times 13 = 364$ different instances of DE. We benchmark the performance of
each DE instance on IOHprofiler~\cite{IOHprofiler}, which contains the~$24$ test functions
from BBOB/COCO~\cite{coco}. We perform experiments on all $24$ test functions, in~$30$ dimensions.
Each DE instance is run~$100$ times on each function, with a function evaluation budget of~$n \cdot
10000 = 300000$, and a population size~$M = 100$~\cite{populationsize}. Due to the computationally
heavy nature of the experiment, it is parallelized using Open MPI~\cite{openmpi} and run on the 
DAS-5 cluster~\cite{das-5}.

To quantify the behavioral impact of BCHMs on DEs, we consider two measures, 1) the best-reached objective function value as a measure of performance and 2) the total percentage of solutions that required repairing or penalization. Since both measures are stochastic, we need to aggregate them through multiple runs, and even over function groups.

We collected the best-reached objective function values from independent runs of each DE variant, and then used these to produce statistical rankings of DE variants, via the well-known the Kolmogorov-Smirnov test\footnote{We used the implementation of DSCTool~\cite{dsctool}.}. To avoid a huge number of pairwise comparisons (which usually drastically decreases the statistical power), we first group the DE variants by their combination of mutation/crossover operators and then compute the ranks for those $13$ variants (differing only in their BCHMs). Furthermore, to make the results more comprehensible, we aggregated the statistical rankings over each function group in BBOB\footnote{$24$ BBOB problems are categorized into five function groups, within which problems share some common characteristics to some degree, e.g., ill-conditioning, regularity, separability,
symmetry, and multimodality.}.
Also, the ranking in each function group is tested for statistical
significance using the Friedman test and the corresponding post-hoc test using the Hochberg procedure, with a
confidence level $\alpha = 0.05$, indicating if a pair of ranks is significantly different.

In addition, we record the percentage of solutions that required a repair or
penalization. For most BCHMs, this percentage is equivalent to the percentage of generated
infeasible donor vectors. The only two exceptions are `Boundary Transformation', where solutions can be repaired even if they are feasible, and `Death Penalty', which is applied to the trial vectors (i.e., after the crossover step).

%% file: results.tex
\input{macros.tex}
\section{Results} \label{sec:results}

\subsection{Comparison of BCHMs} \label{sub:comparison_of_bchms} 
\begin{figure*}[t]
\begin{subfigure}[t]{\textwidth}
\begin{minipage}[c]{0.999\textwidth}
\begin{tabular}{ll}
	\includegraphics[height=3.68cm,trim={3mm 2mm 2mm 3mm},clip]{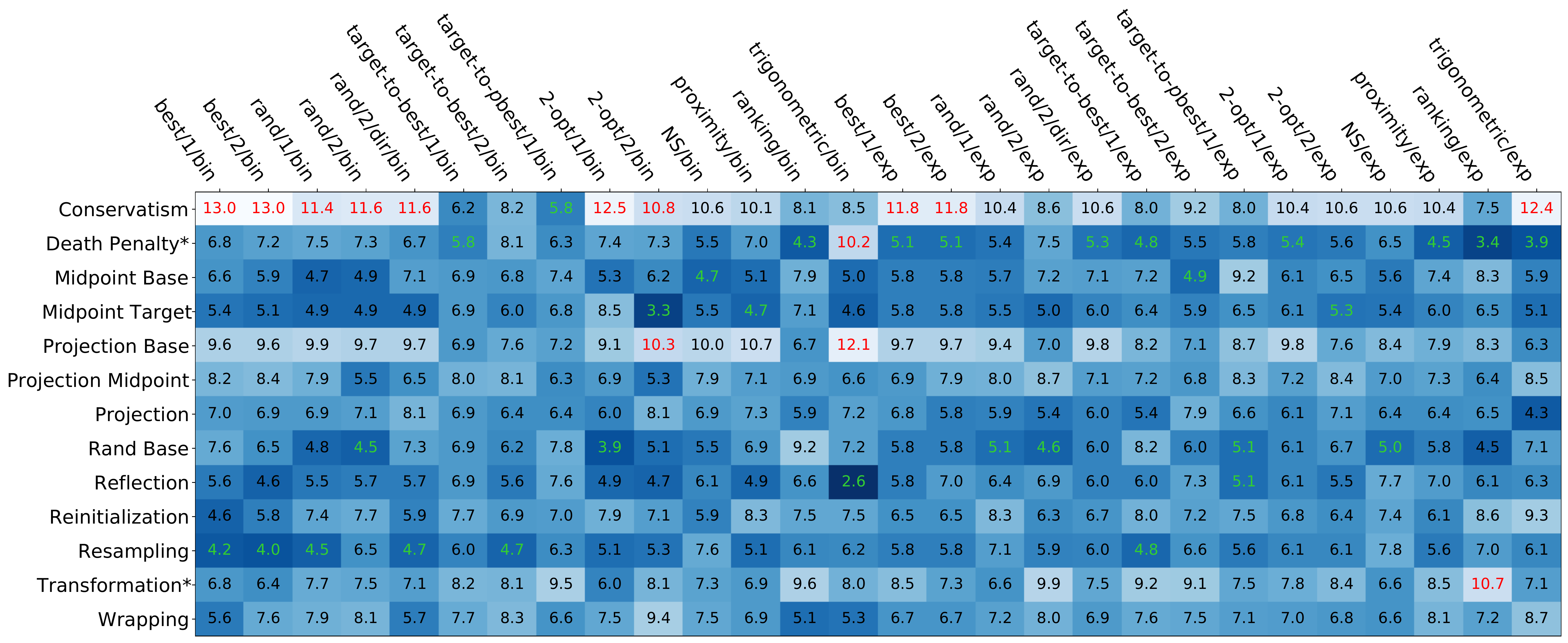} &
	\hspace{-0.4cm}
\includegraphics[height=3.68cm,trim={3mm 2mm 2mm 3mm},clip]{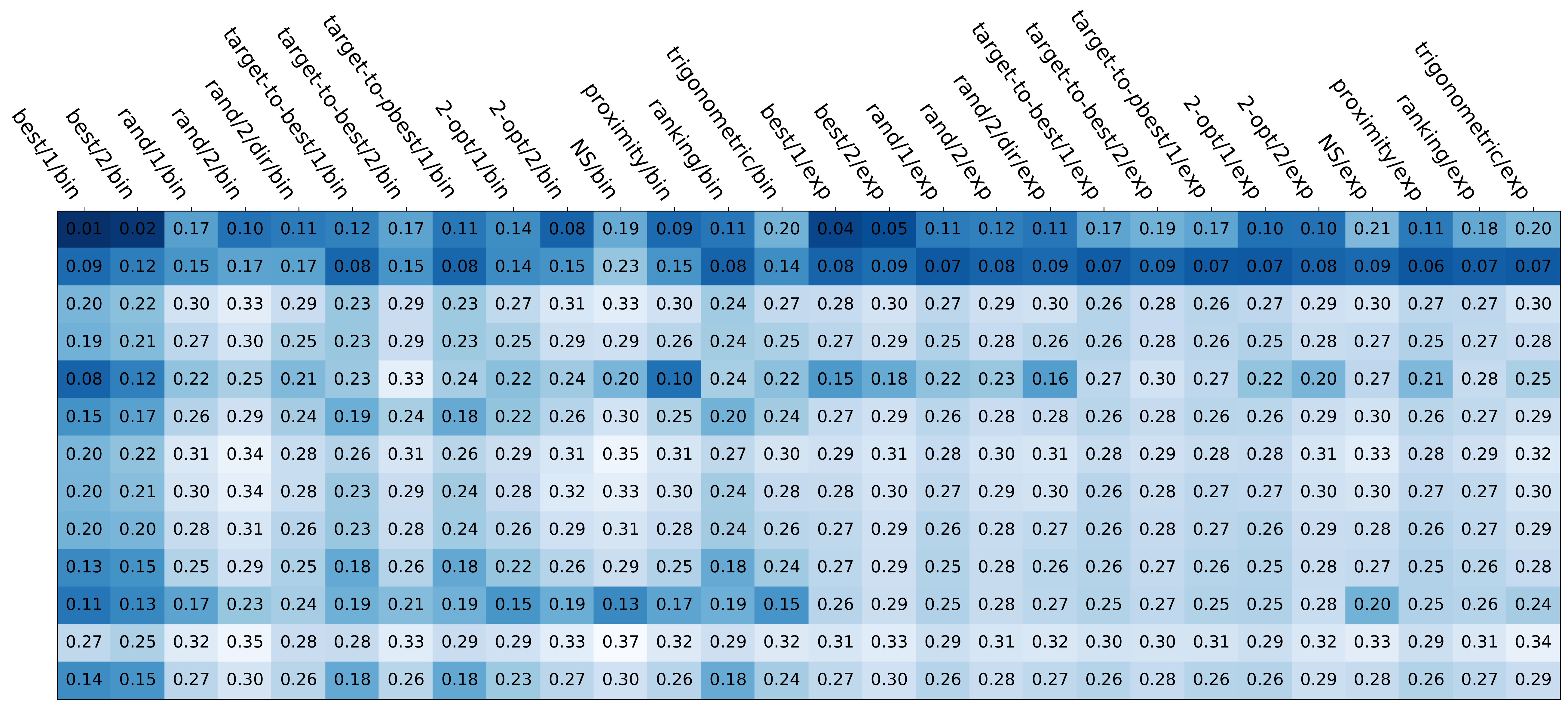} \\
\end{tabular}
\end{minipage}\hfill
\begin{minipage}[c]{0.001\textwidth}
	\hspace{-0.30cm} \rotatebox{270}{\hspace{1cm} \small group 1}
\end{minipage}
\vspace{-0.15cm}
\end{subfigure}
\heatmaprow{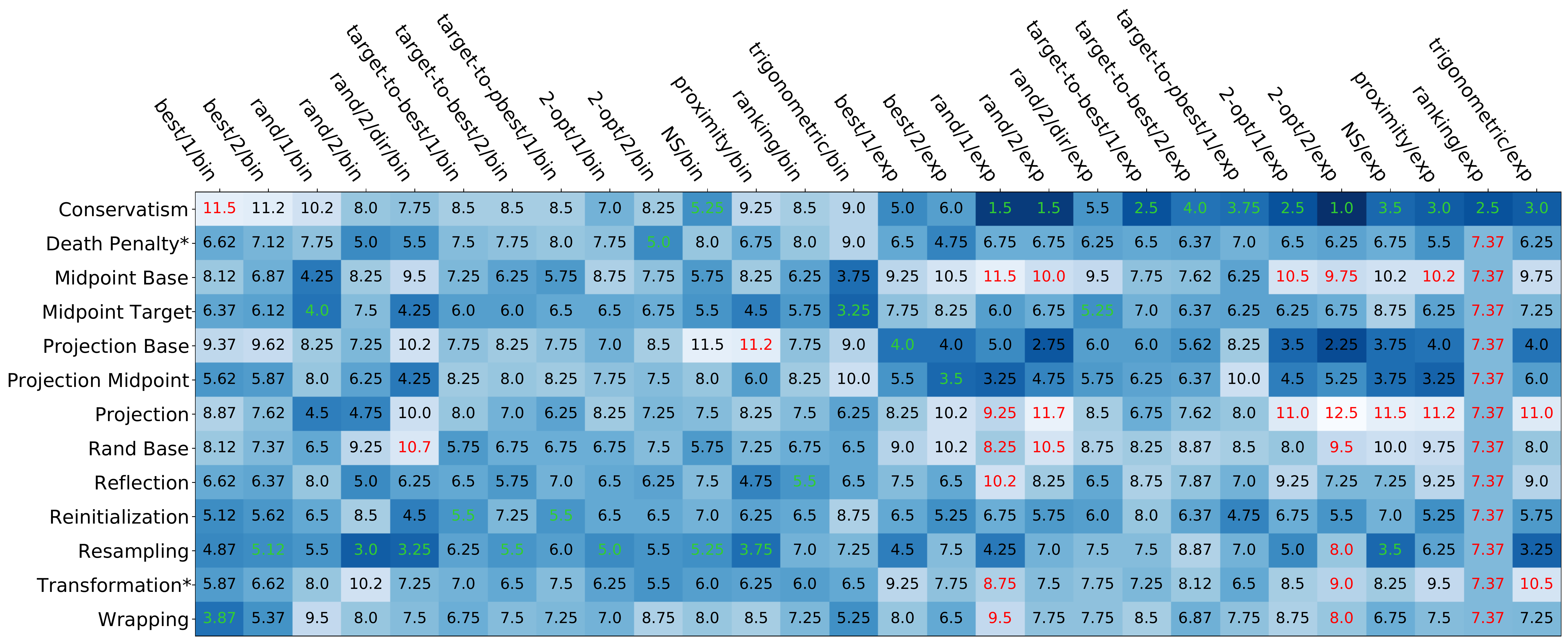}
\heatmaprow{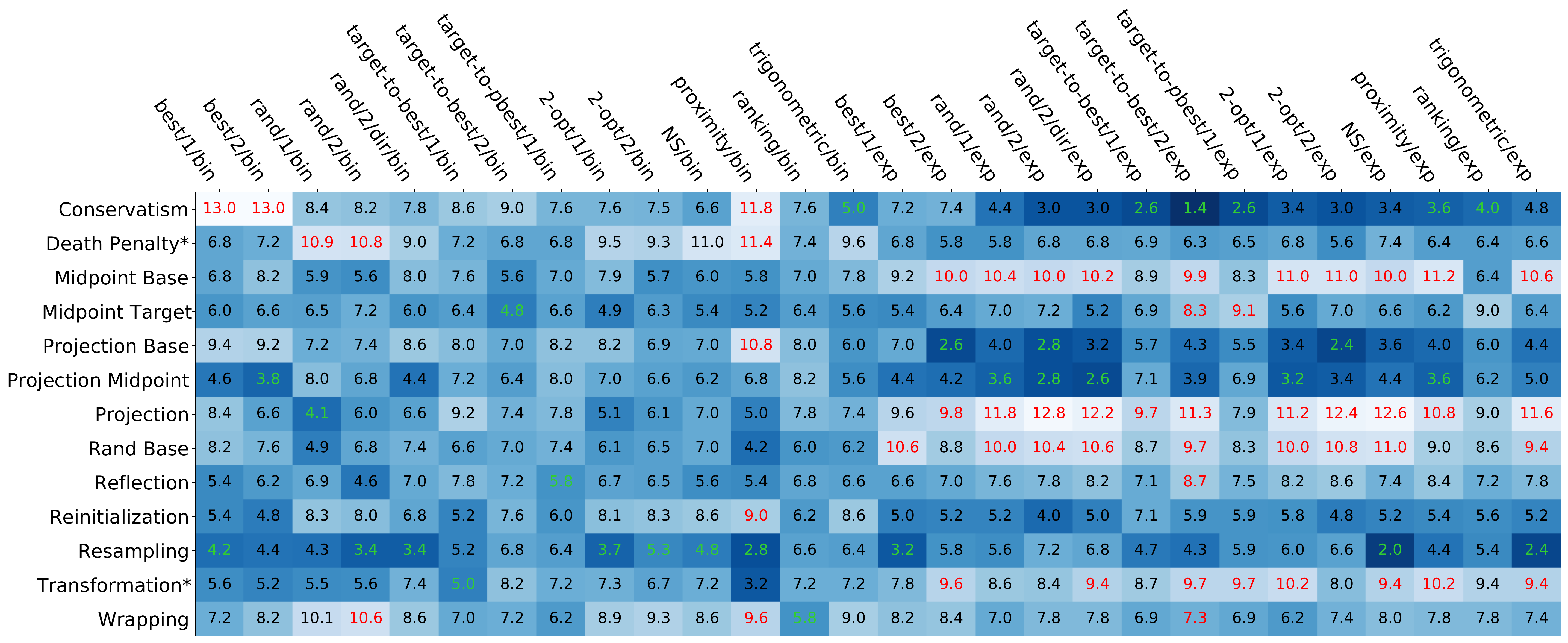}
\heatmaprow{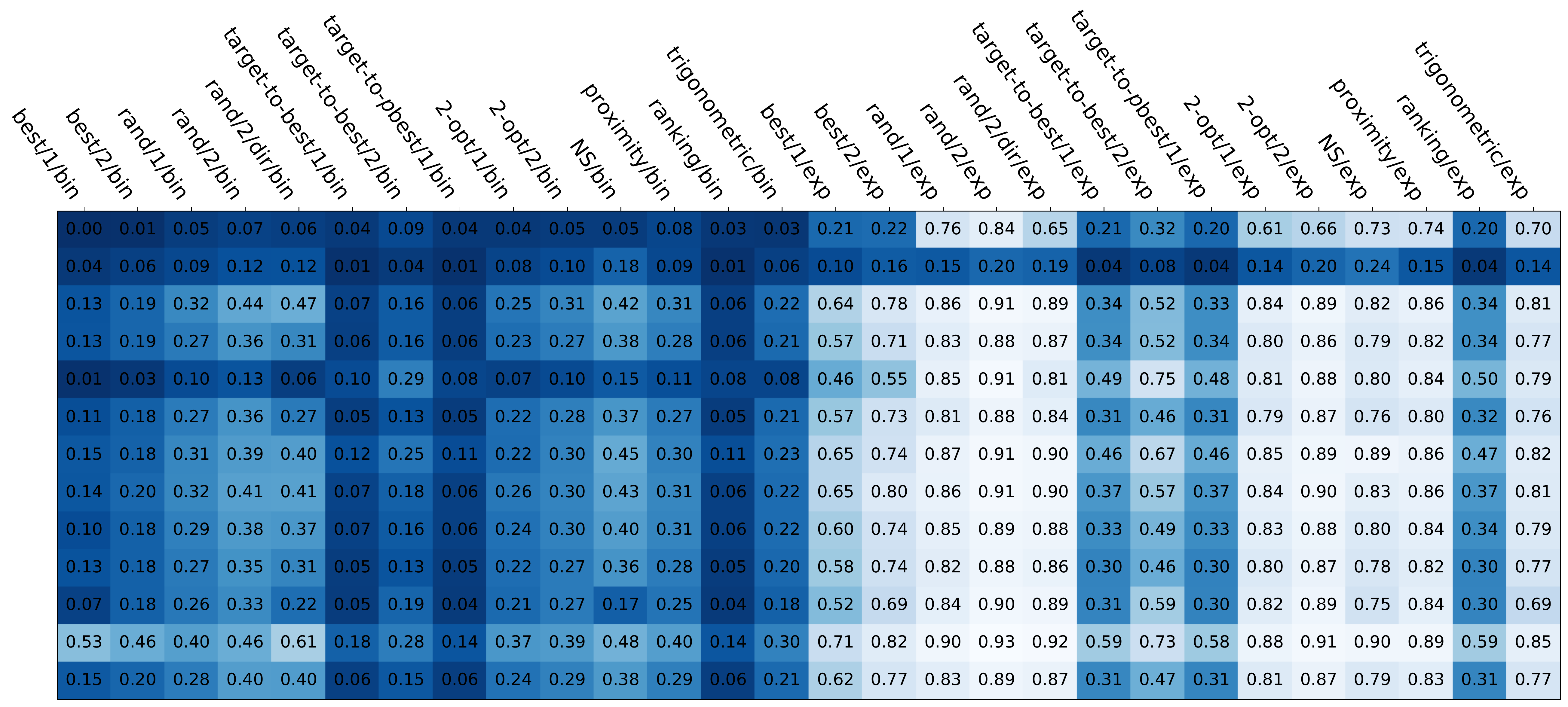}
\heatmaprow{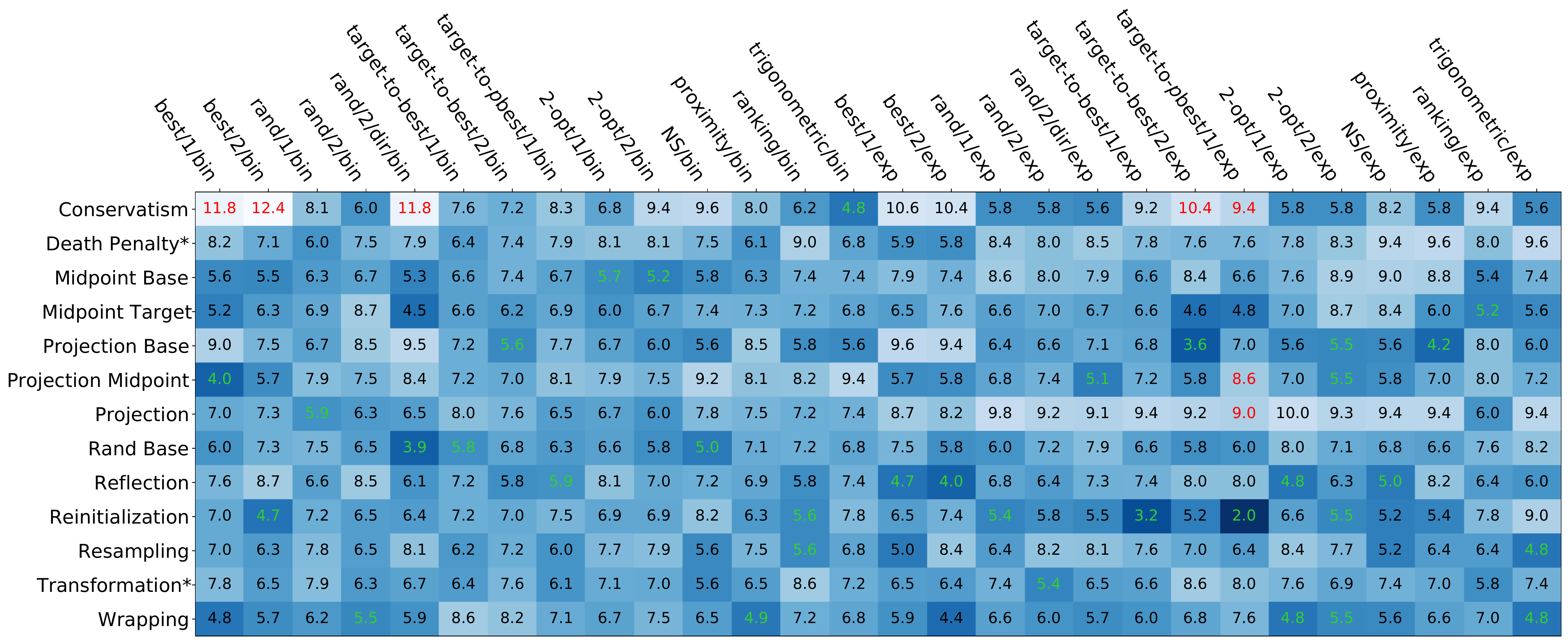}
\caption{Two heatmaps for each BBOB function group. On the left, heatmaps of the mean performance ranks over
of the BCHMs when combined with each mutation/crossover combination. On the right, heatmaps of the
mean percentage of repaired solutions (PORS) of each DE instance. Results are averaged over 100 runs for each function in each function group. See Section ~\ref{sub:comparison_of_bchms} for explanation of cell and font colours.}
\vspace{-0.4cm}
\label{fig:heatmaps}
\end{figure*}

Here, we present the experimental
results. Figure~\ref{fig:heatmaps}, plotted using Matplotlib~\cite{matplotlib}, shows two heatmaps
for each function group. Heatmaps on the left show the mean rank of each BCHM when
combined with a particular mutation and crossover strategy, for the function group in question. Each
cell is colored with a shade of blue, where a darker shade corresponds to a lower (better) rank. The
best ranked BCHM(s) for each mutation/crossover combination in a particular function group are
marked with a green font. All BCHMs that showed significantly worse performance compared to the best
BCHM(s) are marked with a red font. Note that the lowest rank(s) in each column of a heatmap are
always marked green, even if no other BCHM is significantly worse. The heatmaps on the right side of
Figure~\ref{fig:heatmaps} show the average percentage of repaired solutions (PORS) for each DE
configuration combined with each BCHM. To signify the irregularities regarding the Boundary Transformation and Death Penalty BCHMs discussed earlier, they are marked with an asterisk~(*) in Figure~\ref{fig:heatmaps}.

\paragraph{BBOB Function Group 1: Seperable functions (functions 1--5)} 
In the first function group, Death Penalty, Resampling, and Rand Base seem to be a good choice for
most configurations. Conservatism is the worst choice in many cases, and often significantly worse than the best option.  Looking at the
PORS, it is clear that two BCHMs result in the least repairs: Conservatism and Death Penalty. The lower PORS for Death Penalty can in part be due to the fact that it is applied after the crossover step, instead of directly after the mutation step, allowing infeasible components of the donor vector to still be discarded by the crossover operator. The
Transformation BCHM results in the largest PORS, which can likely be explained by its property of altering even feasible solutions. The same can be observed in
the other four function groups. Exploitative DE configurations, for example those
incorporating the `best' vector in their mutation scheme, generate fewer infeasible solutions than exploratory 
ones, resulting in a lower PORS. The PORS in this function group are generally quite high, taking
into account the relatively simple nature of the problems in this function group. This can be
explained by the fact that the optimum of function $5$, `Linear slope', is located on the boundary
of the search space, resulting in PORS of $> 90\%$ for many DE instances.

\paragraph{BBOB Function Group 2: Functions with low or moderate conditioning (functions 6--9)} 
In function group~2, the preferred BCHMs are much more pronounced. In contrast to function group~1,
many configurations now perform best with Conservatism, specifically those using
exponential crossover. For binomial crossover, Resampling is generally a good choice. DE instances seem to be more sensitive to the choice of BCHM when using exponential crossover, as the performance differs with statistical significance more frequently.
The differences in PORS are much larger compared to function group~1. In general, binomial crossover
results in significantly fewer repairs. Many instances using exponential crossover required approximately $50\%$
repairs, perhaps explaining the increased sensitivity to the BCHM choice. However, even with exponential crossover, few repairs are
needed when using the Death Penalty. The `target-to-best' mutation variants generate the fewest
infeasible solutions, with both crossover operators. Note that the `ranking' mutation scheme is a variation on target-to-$p$best/1. 

\paragraph{BBOB Function Group 3: Functions with high conditioning and unimodal (functions 10--14)} 
In function group~3, configurations with exponential crossover generally work best with either
Projection Base, Projection Midpoint, or Conservatism. Resampling is again a good choice for most
configurations with binomial crossover. 
The sensitivity to the BCHM is much higher for DEs with exponential crossover, and especially Midpoint Base, Projection, Rand Base and Transformation often result in significantly worse performance for these DEs. The PORS are similar to those of function group~2.

\paragraph{BBOB Function Group 4: Multi-modal functions with adequate global structure (functions 15--19)} 
The generally preferred BCHMs in function group~4 are comparable to those of function group~3. Midpoint Target additionally performs well in combination with binomial crossover. It seems that the sensitivity to the choice of BCHM is highest in function group 4, especially when using exponential crossover. The optimal choice here is also highly dependent on the mutation strategy (instead of just the crossover strategy). The PORS in
this function group are much higher than in others. Many instances using exponential crossover show over
$80\%$ repairs. We expect these high PORSs to be a result of the high degree of multimodality of
functions $f15 - f19$, stimulating the search of local optima close to the boundary.

\paragraph{BBOB Function Group 5: Multi-modal functions with weak global structure (functions 20--24)}
The differences in terms of performance are much less pronounced in function group~5. It seems
impossible to give a general recommendation for the BCHM here. At the same time, most DE
configurations are less sensitive to the BCHM. Few configurations perform significantly better or worse with different BCHMs. The PORS are overall slightly lower than in function group~4.

\noindent
\paragraph{In general} It is clear that choice of BCHM influences the performance of DE and, thus, should not be disregarded during algorithmic design. Furthermore, choice of BCHM has direct impact on the number of required repairs of infeasible solutions (the so-called PORS). Tracking such number during optimization of the BBOB functions reveals surprisingly high values (up to $93\%$), even for the best performing configurations. The nature of such observation requires further study. DE instances with exponential crossover generate considerably more infeasible solutions, which likely explains their increased sensitivity to the choice of BCHM.

\noindent
\paragraph{Guidelines} \label{par:in_general} Table~\ref{tab:count} shows the number of times each BCHM had the lowest rank for a configuration,
per function group and in total. Conservatism and Resampling are clearly two important BCHMs
to consider, as they perform best with many DE configurations. However, in some cases they are also
poor choices. As a rule of thumb, a practitioner could first try Conservatism for a DE configuration with exponential
crossover and Resampling for one with binomial crossover. Judging from Figure~\ref{fig:heatmaps}, this
policy will, however, not always give optimal results. Therefore, a second option should be considered. For
binomial crossover, Midpoint Target is rarely the optimal choice, but nearly always a \emph{good}
choice. In the exponential crossover case, Projection Midpoint can be employed as a reliable second option.
\begin{table}[t]\tiny
	\centering
	\caption{For each DE configuration, the performance of each BCHM is determined based on their mean ranks
		over a function group. This table shows the number times each BCHM has the best mean rank over
		all such DE configurations. In some cases there were two best BCHMs, in which case a
		column can sum up to more than $28$ (the number of considered DE configurations). The largest value in 
		each column is highlighted in \textbf{bold}.}
	
	\label{tab:count}
	\vspace{-0.5cm}
	\resizebox{\columnwidth}{!}{
	\begin{tabular}{r|rrrrr|r}
		& \multicolumn{5}{c|}{Function group} & \\ \hline
		BCHM & $1$ & $2$ & $3$ & $4$ & $5$ & total \\ \hline
		Conservatism 		& 1 & \textbf{12}& 6 & 4 & 1 & 24 \\
		Death Penalty 		& \textbf{10}& 1 & 0 & 0 & 0 & 11 \\
		Midpoint Base 		& 2 & 0 & 0 & 0 & 2 & 4\\
		Midpoint Target 	& 3 & 3 & 1 & 5 & 1 & 13 \\
		Projection Base 	& 0 & 1 & 3 & 3 & 4 & 11 \\
		Projection Midpoint & 0 & 1 & 6 & 2 & 3 & 12 \\
		Projection 			& 0 & 0 & 1 & 0 & 1 & 2\\
		Rand base 			& 6 & 0 & 0 & 2 & 3 & 11 \\
		Reflection 			& 2 & 1 & 1 & 2 & 5 & 11 \\
		Reinitialization 	& 0 & 2 & 0 & 1 & \textbf{6} & 9 \\
		Resampling 			& 6 & 8 & \textbf{10}& \textbf{12}& 2 & \textbf{38} \\
		Transformation 		& 0 & 0 & 1 & 0 & 1 & 2 \\
		Wrapping 			& 0 & 1 & 1 & 0 & 5 & 7 \\
	\end{tabular}}
\end{table}

\subsection{Comparison of DE Configurations} \label{sec:comparison_of_de_instances}
Since so far we only ranked DE instances with the same mutation and crossover operators, we cannot extract
knowledge about the relative performance between instances with differing mutation and/or crossover. For
this reason, Empirical Cumulative Distribution Functions (ECDFs) with the best DE
instance for each mutation/crossover combination are computed with IOHanalyzer~\cite{IOHprofiler},
and the corresponding graph is plotted with ggplot2~\cite{ggplot}. In Figure~\ref{fig:combined_ecdf}, for each function group, we plot an
ECDF graph with one line for each mutation/crossover combination, where each DE instance uses the best
ranked BCHM for that configuration in the function group in question. This means that the set of
plotted instances differs across the five plots. If there are multiple best BCHMs, the one 
appearing first (from top to bottom) in the heatmaps of Figure~\ref{fig:heatmaps},
is used.  The ECDF shows the average proportion of targets hit across all runs in a certain function group on the $y$-axis, given a number of
used function evaluations, which is displayed on the $x$-axis. The number of function evaluations is
divided by $30$, the dimensionality of the benchmark problems. The targets are $f_{\text{opt}} +
\{10^1, \dots ,10^{-8}\}$, where $f_{opt}$ is the objective function value of the optimum. The
mutation scheme is encoded using color, and the crossover by the line type, where a solid line
indicates binomial crossover and a dashed line exponential crossover.
\paragraph{BBOB Function Group 1}
In function group 1, configurations with exponential crossover perform
significantly better than those with binomial crossover. In fact, all instances with
exponential crossover, except the one using best/1 mutation, reach $100\%$ of the targets within the allocated budget, while only one instance with binomial crossover (using target-to-best/2 mutation) does. Best/1 and best/2 mutations combined with
binomial crossover show to be too exploitative, as they hit many targets quickly but seem get stuck
in local optima, preventing them from hitting the final target.
\paragraph{BBOB Function Group 2}
Interestingly, binomial crossover
generally performs better than exponential in function group 2. A large portion of the instances using exponential crossover reaches fewer than $10\%$ of the targets. All target-to-best variants, in
particular target-to-best/2 and ranking-based mutation, perform very well when combined with
binomial crossover. Ranking-based mutation is a good choice regardless of the crossover operator. In fact, the
relative performances of the mutation schemes is very similar between the two crossover schemes. 
\paragraph{BBOB Function Group 3}
The ECDF of function group 3 is very similar to that of function group 2, but fewer targets are reached by most instances. The `target-to-best'
mutation schemes are again most successful. The performance difference between binomial and
exponential crossover is more pronounced in this function group; all instances using binomial
crossover perform better than all instances using exponential crossover. 
\paragraph{BBOB Function Group 4}
Similar to function groups~2 and~3, binomial crossover outperforms exponential crossover in function
group 4.  The best mutation schemes, are,
however, completely different. Each of the top 5 DE instances (all of which use binomial crossover) 
in this function group use an exploratory mutation scheme: rand/1, rand/2, or variations thereof:
2-opt/1, 2-opt/2 and proximity-based rand/1. In contrast, the instances with exponential crossover
performed better with exploitative mutation schemes. 
\paragraph{BBOB Function Group 5}
As in function group~1, DE instances with exponential crossover outperform most instances using 
binomial crossover in function group 5. The best instance used the NSDE mutation scheme. Furthermore, the 
best/1 and all the `target-to-best' mutation schemes performed well. 
\paragraph{In general -- guidelines} 
The ECDFs show an advantage for exponential crossover in function groups $1$ and $5$,
one for binomial crossover in function groups $2$, $3$ and $4$. The performance
difference between instances differing only in the crossover operator can be huge. Therefore, it is
especially important to carefully consider the choice of crossover operator based on the characteristics
of the problem at hand. In all function groups except function group $4$, DE instances using
binomial crossover perform best with `target-to-best' mutation variants. In function group $4$,
exploratory mutation schemes like rand/1 or 2-opt/1 performed better. This can be explained by the
high degree of multimodality of the test functions in this function group, where exploratory mutation
schemes are more likely to escape local optima. When using exponential crossover, too, a `target-to-best'
mutation variant is often a good choice, as well as best/1 mutation or the neighborhood search
mutation operator from NSDE. It is important to note that the choice of BCHM, in this case the optimal choice according to Figure~\ref{fig:heatmaps}, can, as demonstrated, have a significant impact on the performance of the DE configurations, and different results could be obtained by selecting different BCHMs.

%% file: macros.tex
\newcommand{\heatmaprow}[1]{
  \begin{subfigure}[t]{\textwidth}
  \begin{minipage}[c]{0.995\textwidth}
  \begin{tabular}{ll}
	  \includegraphics[height=2.6075cm,trim={3mm 2mm 2mm 7.12cm},clip]{heatmap_perf_group#1.pdf} &
	  \hspace{-0.4cm}
  \includegraphics[height=2.6075cm,trim={3mm 2mm 2mm 7.12cm},clip]{heatmap_perc_group#1.pdf} \\
  \end{tabular}
  \end{minipage}\hfill
  \begin{minipage}[c]{0.005\textwidth}
	  \hspace{-0.22cm} \rotatebox{270}{\centering \small group #1}
  \end{minipage}
  \vspace{-0.15cm}
  \end{subfigure}
}

%% file: conclusion.tex
\section{Conclusions and Future Work} \label{sec:conclusion}
We aim to quantify the impact of boundary constraint
handling methods (BCHMs) on Differential Evolution (DE) algorithms in terms of the empirical performance and algorithm's behavior. For this purpose, we took the so-called modular DE framework~\cite{PSODE}, which is capable of instantiating a huge number of DE variants by combining different mutation, crossover, and BCHM operators.

This paper puts a special emphasis on the BCHM, as this operator is often
overlooked in existing literature. In detail, $14$ mutation operators, $2$ crossover operators,
and $13$ BCHMs have been tested in this work, resulting in $364$ DE instances. We benchmarked those instances on the well-known BBOB/COCO problem set~\cite{coco} in $30$ dimensions. The experimental results
were aggregated over each of the five function groups, where the members of each function group
share similar characteristics. 

As for the empirical performance, we measure the best-reached function value. The results show that the choice of BCHMs requires careful consideration, as it can impact the performance of DE significantly. The best choice depends on the problem to optimize
and the DE instance to use, but general guidelines can be given based on the crossover
operator, as this seems to have by far the greatest influence on the optimal choice. For a DE instance using exponential 
crossover, Conservatism is the best choice for BCHMs concerning most mutation
operators and most function groups, but it can result in sub-optimal results in some
cases, where using Projection Midpoint as a second option would improve the performance. For the
binomial crossover, we recommend to employ Resampling as the initial choice, 
and consider Midpoint Target as a reliable fallback option.

For quantifying the algorithm's behaviour with respect to boundary constraints, we recorded the percentage of repaired solutions (PORS) and observed that a large difference of PORS between DE instances
using binomial crossover and those using exponential crossover, which is likely explained by the
more aggressive perturbations made by the exponential variant. We expect the higher PORS to be the reason that DEs with exponential crossover are much more sensitive to the choice of BCHM, compared to those using binomial crossover. As a high PORS indicates a higher
level of exploration near the boundary of the search space, it could be beneficial to switch from
exponential to binomial crossover during the course of the run, favoring exploitation in the later
stages of the optimization process. The `target-to-best' mutation variants generally generate fewer
infeasible solutions than others. Additionally, these mutation schemes showed to perform well
in most function groups. The BCHM has also been shown to have a direct impact on the PORS. We have, however, not been able to show causality between the PORS and the
performance of a DE instance. This is an interesting direction for future efforts. 

In our experiments, we observed a significant difference in performance between instances with
different crossover operators. Instances using exponential crossover performed better in function
groups 1 and 5, and those with binomial crossover performed better in function groups 2, 3, and 4.
The four `target-to-best' mutation scheme variants we experimented with showed the best performance
overall. In some cases, a more exploratory mutation scheme like rand/1 or NSDE yielded better
results. 

In future work, the experiments should be repeated in more dimensionalities, as the results could
vary. We expect the BCHM to be even more critical in higher dimensionalities, where more infeasible solutions are generated~\cite{outside-the-box}. Further, the large number of DE operators present in the implemented software framework arouse the interest
for adaptive selection of operators, similar to SaDE~\cite{sade}. This possibility should also be
explored in future efforts.